\documentclass[letterpaper, 10 pt, conference]{ieeeconf} 
\IEEEoverridecommandlockouts
\usepackage{cite}
\usepackage[font=footnotesize]{subfig}
\usepackage{multirow}
\usepackage{ifpdf}
\usepackage[pdftex]{graphicx}
\usepackage{epstopdf}
\usepackage[cmex10]{amsmath}
\usepackage{algorithmic}
\usepackage{array}
\usepackage{mdwmath}
\usepackage{mdwtab}
\usepackage{eqparbox}
\usepackage{verbatim}

\usepackage[font=footnotesize]{subfig}
\usepackage{fixltx2e}
\usepackage{stfloats}

\usepackage{amsmath,bm}
\usepackage{amssymb}
\usepackage{color}
\usepackage{subfig}

\usepackage{multirow}
\usepackage{float}
\usepackage{caption}
\usepackage{comment}

\usepackage{makecell}
\usepackage{amsmath}
\usepackage{graphicx}
\usepackage[colorlinks=true, allcolors=blue]{hyperref}
\usepackage{adjustbox}
\usepackage{blindtext}
\usepackage{hyperref}

\usepackage[font=footnotesize]{subfig}
\usepackage{fixltx2e}
\usepackage{stfloats}

\usepackage{amsmath,bm}
\usepackage{amssymb}
\usepackage{color}
\usepackage{subfig}

\usepackage{multirow}
\usepackage{float}
\usepackage{caption}
\usepackage[normalem]{ulem}
\usepackage{comment}
\usepackage{hyperref}
\usepackage{cite}
\usepackage[font=footnotesize]{subfig}

\usepackage{ifpdf}

\usepackage{epstopdf}
\usepackage[cmex10]{amsmath}

\usepackage{array}
\usepackage{mdwmath}
\usepackage{mdwtab}
\usepackage{eqparbox}
\usepackage{verbatim}
\usepackage{booktabs}
\usepackage{multirow}
\usepackage{graphicx}
\usepackage{algorithm}
\usepackage{algorithmic}
\usepackage{array} 
\usepackage[normalem]{ulem}  
\usepackage{booktabs}

\begin{document}
	\title{\LARGE \bf
Deep Reinforcement Learning-Based Semi-Autonomous Control for
Magnetic Micro-robot Navigation with Immersive Manipulation
}

\author{
Yudong Mao, Dandan Zhang
\thanks{Yudong Mao, Dandan Zhang are with the Department of Bioengineering, Imperial-X Initiative, Imperial College London, London, United Kingdom.  Corresponding: d.zhang17@imperial.ac.uk.}
}

\maketitle
\begin{abstract}
Magnetic micro-robots have demonstrated immense potential in biomedical applications, such as in vivo drug delivery, non-invasive diagnostics, and cell-based therapies, owing to their precise maneuverability and small size. However, current micromanipulation techniques often rely solely on a two-dimensional (2D) microscopic view as sensory feedback, while traditional control interfaces do not provide an intuitive manner for operators to manipulate micro-robots. These limitations increase the cognitive load on operators, who must interpret limited feedback and translate it into effective control actions.
To address these challenges, we propose a Deep Reinforcement Learning-Based Semi-Autonomous Control (DRL-SC) framework for magnetic micro-robot navigation in a simulated microvascular system. Our framework integrates Mixed Reality (MR) to facilitate immersive manipulation of micro-robots, thereby enhancing situational awareness and control precision. Simulation and experimental results demonstrate that our approach significantly improves navigation efficiency, reduces control errors, and enhances the overall robustness of the system in simulated microvascular environments.

\end{abstract}



\section{Introduction}

Over the past few decades, significant advancements have been made in the manipulation of magnetic micro-robots, facilitating precise drug delivery \cite{luo2018micro,lin2024magnetic,vikram2016targeted}, non-invasive diagnostics \cite{zhang2023advanced}, and cell-based therapies \cite{wang2021trends, zhang2022fabrication, wang2023microrobots}. Magnetic micro-robots are typically small, enabling them to navigate complex and narrow spaces within the human body in a non-contact manner \cite{wang2021endoscopy}. These micro-robots can be driven using various systems, including permanent magnets \cite{abbes2019permanent, nadour2023cochlerob}, electromagnetic arrays \cite{lee2018capsule, lee2020needle}, and Helmholtz or Maxwell coil systems \cite{jeong2020acoustic, gong2024magnetic, abdelaziz2022electromagnetic}. While electromagnetic arrays or coil systems offer simpler designs and more uniform magnetic fields, they are limited by smaller working space volumes and higher energy consumption \cite{jeong2020acoustic}. Permanent magnets can generate strong localized magnetic fields and can facilitate more flexible control in complex environments \cite{erni2013comparison}.

Despite these technological advancements, most practical therapeutic applications, such as drug delivery, still rely heavily on teleoperation for manual control of magnetic micro-robots \cite{lee2021real}. Traditional teleoperation systems provide operators with only 2D real-time operational images, resulting in less intuitive visual feedback \cite{Jang_2019}. These limitations lead to insufficient perception of the 3D microscale environment by the operator, making precise and efficient control of micro-robots challenging. \cite{mehrtash2011bilateral, khatib2020teleoperation}.

\begin{figure}[t]
  \captionsetup{font=footnotesize,labelsep=period}
    \centering
    \includegraphics[width=1.05\linewidth]{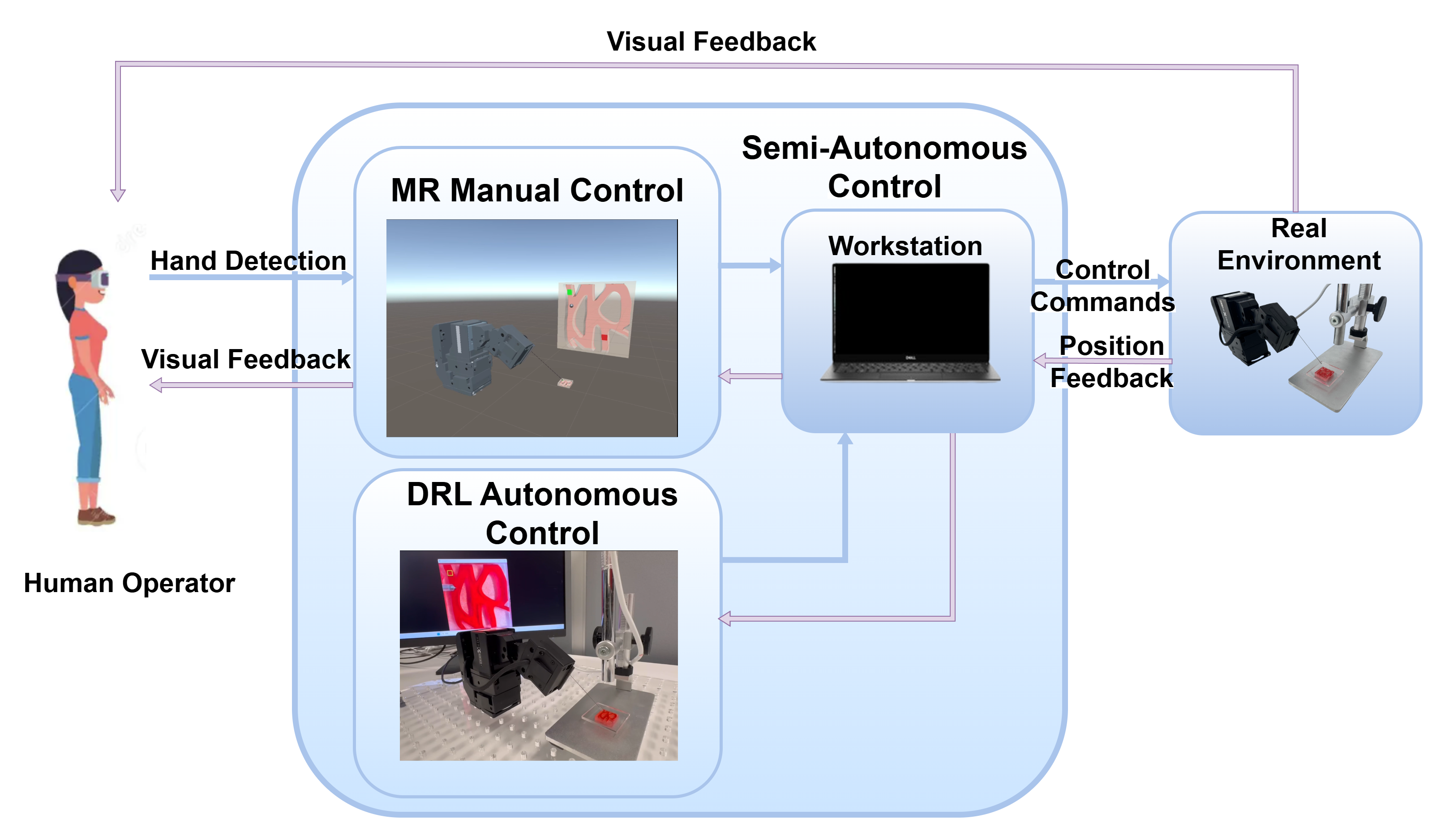} 
    \caption{Schematic of the Deep Reinforcement Learning-Based Semi-Autonomous Control (DRL-SC) framework for magnetic micro-robot navigation with Mixed Reality (MR)-based immersive manipulation. The diagram illustrates the key components of the framework, including MR-based manual control, DRL-based autonomous control, and the physical environment for magnetic micro-robot navigation.}
    \label{fig:concept}
     \vspace{-0.6cm}
\end{figure}

Mixed Reality (MR) is a technology that integrates information between the physical and digital worlds, offering operators an immersive experience \cite{jiang2024digital, hoenig2015mixed, fan2023digital}. MR technology can provide crucial information required during teleoperation, including 3D visualization of the objects being manipulated and enable interaction with real-world objects, thereby enhancing the safety and effectiveness of the operations \cite{zhang2024hubotverse}. For example, Chowdhury et al. combined MR with an electromagnetic drive system, enabling operators to intuitively manipulate micro-robots within 3D-printed vascular models, which offers a more immersive operational experience \cite{chowdhury2024virtual}. Therefore, we aim to integrate MR into magnetic micro-robotic system to facilitate immersive and intuitive micromanipulation.


Another significant challenge for magnetic micro-robots is achieving intelligent and precise navigation \cite{salehi2024intelligent}. Traditional micro-robot navigation typically involves mathematically modeling the micro-robot's operational environment to derive a dynamic model for navigation \cite{li2024automated}.
However, the highly dynamic and complex nature of the operational environment of magnetic micro-robots makes traditional dynamic modeling both difficult and unreliable \cite{das2024multifaceted}.

To address the challenges mentioned above, Deep Reinforcement Learning (DRL) has been investigated as a solution for micro-robot navigation in unstructured environments, which is suitable for the magnetic micromanipulation tasks \cite{wang2024deep}. Salehi et al. utilized the Soft Actor-Critic (SAC) and Trust Region Policy Optimization (TRPO) algorithms to control a disk-shaped magnetic micro-robot on a water surface \cite{salehi2024intelligent}.
Abbasi et al. integrated DRL controllers with A*/D* algorithms for 3D navigation in cerebral vessels. They developed an electromagnetic drive system in Unity and training a controller using Proximal Policy Optimization (PPO), which improved navigation accuracy and reduced training time \cite{abbasi2024autonomous}.
Behrens et al. created helical magnetic hydrogel micro-robots controlled by non-uniform, nonlinear, time-varying magnetic fields via a tri-axial coil array. They used images captured from the environment as input for the SAC algorithm and defined the action space by coil currents, achieving end-to-end DRL navigation \cite{behrens2022smart}.  While DRL-based autonomous control reduces the workload for operators during micromanipulation, human-in-the-loop remains crucial, particularly for biomedical applications. Approaches such as semi-autonomous control and human-robot shared control are worth investigating to enhance safety and reliability in these biomedical applications \cite{zhang2022human}.



To this end, we aim to develop a semi-autonomous control approach, which integrates the advantages of human control with robotic precision and autonomous motion generation \cite{9341383}. To the best of our knowledge, this is the first DRL-based semi-autonomous control framework (DRL-SC) with MR-driven immersive manipulation for the navigation of magnetic micro-robots in biomedical settings, as shown in Fig.\ref{fig:concept}. Our approach integrates the learning capabilities of DRL with human supervision. In this semi-autonomous setup, the human operator can take over control in critical situations, ensuring a safe and effective overall navigation process.
This framework not only enhances operators' control accuracy but also reduces their workload.

The main contributions are as follows:

\begin{itemize}
    \item A digital twin-driven MR-based immersive manipulation system was constructed for magnetic micro-robot manipulation, which enhances the ergonomics of control.
    \item A DRL-SC framework was proposed, which combines the advantages of DRL-based autonomous control and human operator decision-making to ensure safety for potential biomedical applications.
    \item Three simulated microrobot navigation control tasks were performed in 3D-printed vascular models, followed by a user study comparing performance with and without DRL-based semi-autonomous control.
\end{itemize}

\section{Methodology}

\subsection{Hardware System} 

\begin{figure}
    \captionsetup{font=footnotesize,labelsep=period}
    \centering   \includegraphics[width=\linewidth]{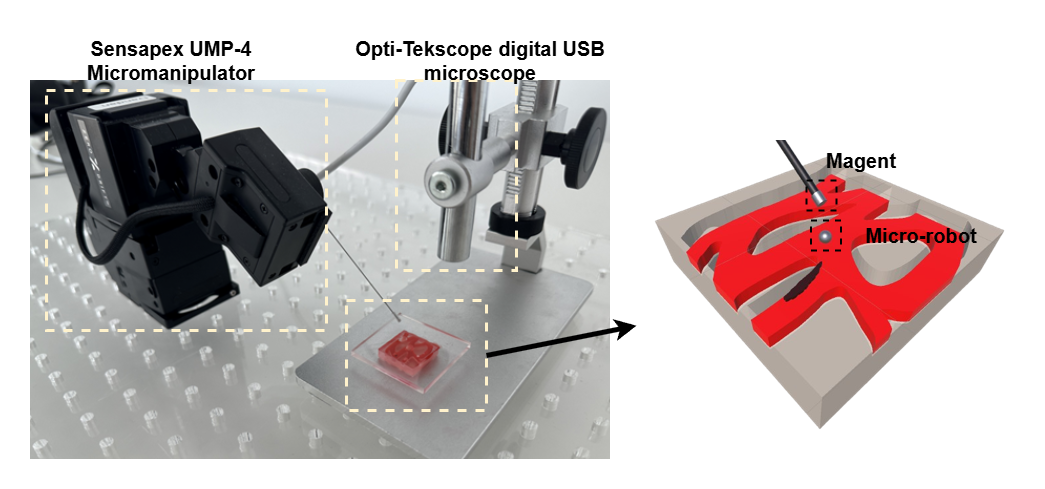} 
    \caption{Overview of the experimental setup: This system includes a micromanipulation with a magnetic microneedle for non-contact manipulation of the magnetic micro-robot, a digital microscope for acquiring high-resolution 2D microscopic views, and a magnified visualization of the simulated microvascular environment.}
    \label{fig:example_magnet}
            \vspace{-0.6cm}
\end{figure}

The system, illustrated in Fig.\ref{fig:example_magnet}, comprises a high-precision UMP-4 micromanipulator (Sensapex Oy). The micromanipulator features four independent motion axes. The X, Y, and Z axes correspond to three-dimensional translational motion. The W-axis is dedicated to micro-tool insertion, allowing controlled linear movement along a specific direction. These axes have a motion range of 0 to 20,000 micrometers. An adjustable-length iron needle, is mounted at the end of the micromanipulator. A cylindrical permanent magnet with a diameter of 1 mm, made of neodymium-iron-boron (NdFeB) material, is attached to the tip of the needle. This magnet exhibits a very strong localized magnetic field, with the magnetic flux density measured at approximately 100 mT. 
To ensure that the permanent magnet on the needle does not adsorb the micro-robot, a critical distance was set. This distance is determined through measurement and by calculating the balance between magnetic force and gravity. An Opti-Tekscope digital USB microscope (1600x1200 pixels, 30 frames per second) is used during operation to capture top-down 2D images as real-time visual feedback. The magnetic micro-robot is a spherical permanent magnet with a diameter of 1 mm. When the cylindrical permanent magnet is moved under the control of the micromanipulator, it generates magnetic field to actuate the magnetic micro-robot in a non-contact manner.
The magnetic force acting on the micro-robot is given by:
\begin{equation}
\mathbf{F} = \left( \mathbf{M} \cdot \nabla \right) \mathbf{B}
\label{eq:force}
\end{equation}
where \( \mathbf{F} \) is the magnetic force on the micro-robot,
\( \mathbf{M} \) is the magnetic moment of the micro-robot,
\( \mathbf{B} \) is the external magnetic field,
\( \nabla \) represents the gradient operator.

\subsection{Immersive Teleoperation Framework with Digital Twin} 

To provide operators with enhanced visual feedback and a more intuitive micromanipulation experience, this study proposes an immersive framework with DT. This framework primarily consists of a forward path for transmitting control information and a feedback path for visual feedback. 
The HoloLens 2 (Microsoft) uses an external camera to detect the operator’s hand positions and transmits this position data to the micromanipulator at a rate of 60 updates per second, in real-time, via the TCP communication protocol. Incremental control is used to convert hand position increments into micromanipulator displacements, similar to the work in \cite{zhang2020microsurgical}.  To prevent operator tremors, the upper and lower limits of increments are set to limit unexpected displacements. 
The end position of the micromanipulator and the real-time position of the micro-robot are transmitted back to the HoloLens 2 via TCP \cite{jiang2024adaptive}. Based on the real-time position information obtained from the actual environment, a precise and reliable DT of the system can be constructed \cite{fan2023digital}. Once the system's DT is established, it is deployed into the virtual scene of the HoloLens 2. At this point, the operator can observe not only the system's DT within the HoloLens 2 but also the real-time 2D images of the experimental setup captured by the microscope through the lenses.



\subsection{DRL-Based Autonomous Control} 
\subsubsection{DRL Environment Definition and Reward Function}
In this study, DRL is initially used under simplified conditions, where state information for the micro-robot system is derived by creating image-processing-based state vector inputs. The learning process in RL is based on interactions with the environment over a series of time steps to learn optimal policies that maximize the expected cumulative reward, as shown in Equation (2).
This process can be described as a Markov Decision Process (MDP). An MDP is typically represented by a tuple ($S$, $A$, $P$, $R$, $\gamma$), where $S$ is the state space, $A$ is the action space, $P$ is the state transition matrix, $R$ is the reward function, and $\gamma$ is the discount factor.

\begin{equation}
\sum_{t=1}^{T} E_{\left(s_t, a_t\right) \sim \rho_\pi} \left[r_t\right]
\label{eq:reward}
\end{equation}
where $T$ is the total number of time steps and $\rho_\pi$ represents the discounted state-action visitation.

Due to the inherent trial-and-error nature of RL, the cost of directly training in real-world environments is extremely high. Moreover, during training, unavoidable scenarios may arise (e.g., the micro-robot getting stuck in a certain area), making environment reset challenging. Training in a simulated environment reduces the difficulty of resetting the environment and saves time during the learning process.

The interactive environment for RL model training is defined as a grid world with dimensions of 1800x1800 pixels. It is designed to align with the structure of the 2D vascular model. The agent is represented as a circular entity with a radius of 50 pixels. Within this grid world, the regions corresponding to the vascular pathways are designated as navigable, while the remaining areas are considered non-navigable. At the beginning of each episode, both a starting position and a target position are randomly selected from the navigable regions, and the agent's position is initialized at the starting location. The episode terminates either when the agent successfully reaches the target position or when the number of time steps exceeds 20,000.

The state space comprises six elements, including the current 2D position of the micro-robot $(x, y)$, the initial release position $(x_{\text{start}}, y_{\text{start}})$, and the final target position  $(x_{\text{target}}, y_{\text{target}})$. 
The action space consists of 40 discrete action pairs, allowing the robot to choose any integer coordinate position on the square outline with a center at $(0,0)$ and a side length of 10 as the displacement vector for its actions, which facilitates smooth agent movement and makes it more possible to find the optimal path.
The reward function consists of three parts, defined as follows:
\begin{equation}
\text{reward} = 
\begin{cases} 
1000, & \text{arrive} \\
-10, & \text{hit the wall} \\
-0.005 \times d - 0.02 \times t, & d > 10 
\end{cases}
\label{eq:reward_formula}
\end{equation}
where $d$ is the Euclidean distance between the agent and the target position in pixels, and $t$ is the magnitude of the displacement vector for each action taken by the agent, measured in pixels.

If the Euclidean distance between the agent and the target position is less than 10 pixels, the episode ends with a substantial reward, significantly encouraging the agent to reach the endpoint. If the agent’s action would result in reaching an infeasible area, it is considered a wall hit, the position remains unchanged, and a -10 penalty is assigned to reduce the likelihood of taking that action near the edges. A negative reward based on the Euclidean distance to the target position guides the agent toward the goal and penalizes the agent to minimize the number of time steps required to reach the target. Lastly, a penalty associated with the displacement vector is set, making each action non-equivalent.  The weights and selection of sub-rewards are determined based on experience to induce the micro-robot to reach the target position without any unexpected behavior. The reward at each time step ranges from -23 to 1000.

\subsubsection{Implementation of PPO and A2C Algorithms}


\begin{figure}
  \captionsetup{font=footnotesize,labelsep=period}
    \centering
    \includegraphics[width=\linewidth]{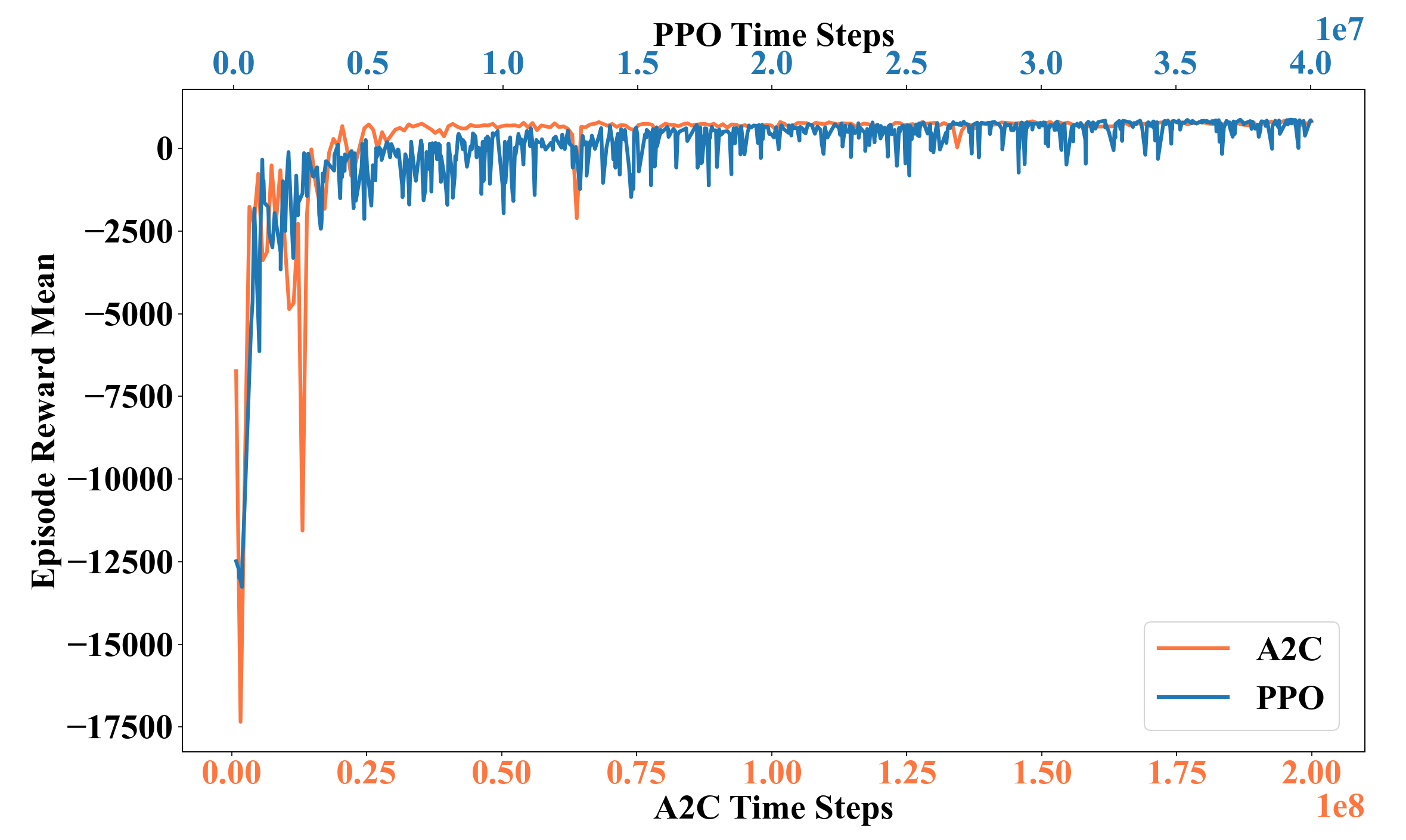} 
    \caption{The model learning curve of PPO and A2C.}
        \vspace{-0.6cm}
    \label{fig:example-curve}
\end{figure}

Since the task involves a discrete action space, this study compares two DRL algorithms, Proximal Policy Optimization (PPO) and Advantage Actor-Critic (A2C), for the autonomous navigation of micro-robots \cite{mnih2016asynchronous,schulman2017proximalpolicyoptimizationalgorithms}.
The difference in the composition of their loss functions leads to significant differences in the stability of policy updates, sample efficiency, and convergence speed when training the policy networks. As shown in the comparison (Fig. \ref{fig:example-curve}), PPO exhibits superior sample efficiency and more stable policy updates, allowing it to reach convergence in fewer time steps. This rapid convergence, combined with stable policy updates, makes PPO particularly well-suited for applications that require deployment in real-world environments. Consequently, this study selected the PPO algorithm to train the autonomous navigation strategy for the magnetic micro-robots.

\begin{figure}
    \captionsetup{font=footnotesize,labelsep=period}
    \centering
    \includegraphics[width=\linewidth]{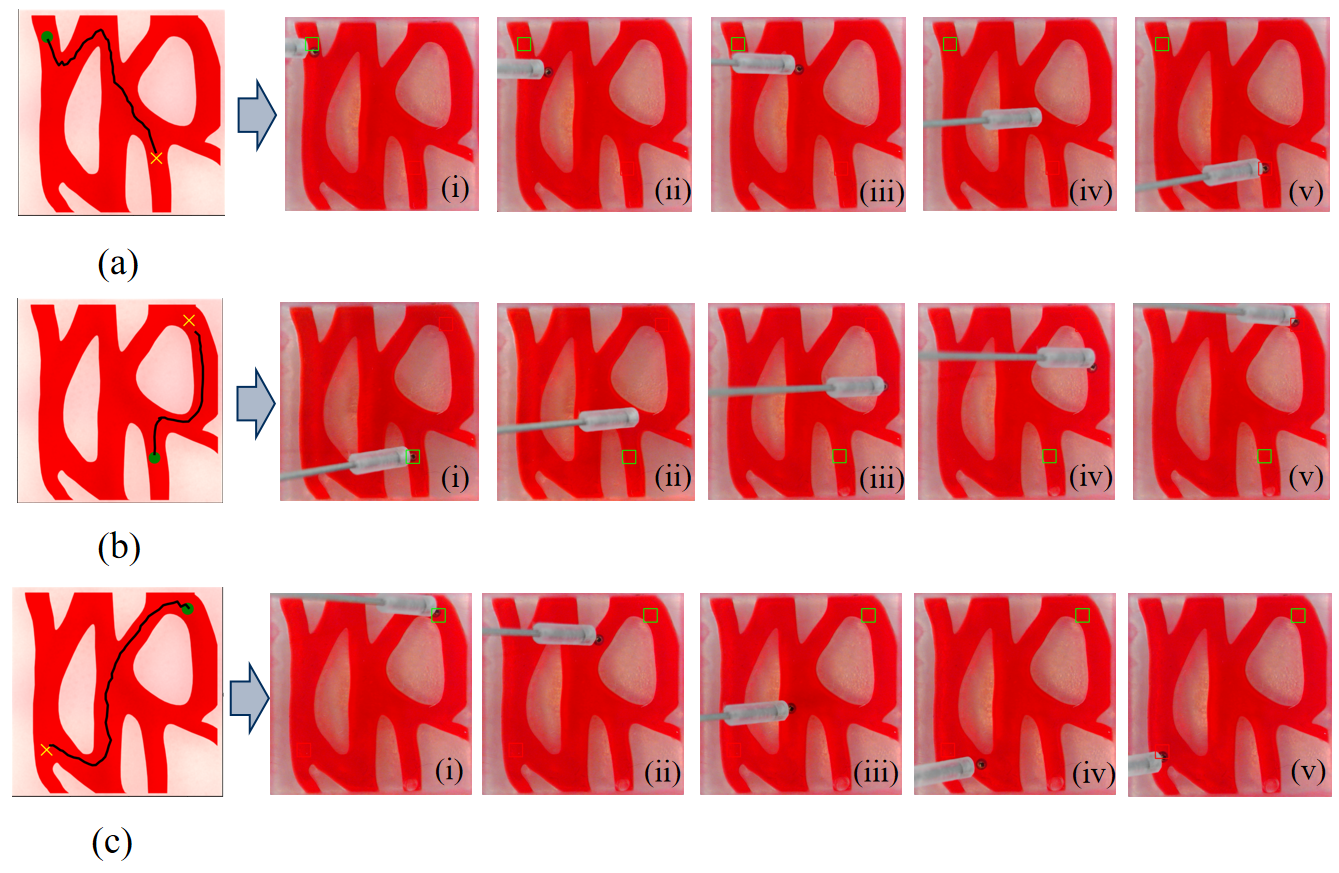} 
    \caption{ (a), (b), and (c) each show three sets of experiments with randomly selected starting and ending points, illustrating the navigation performance of the PPO model in both virtual and real environments.}
    \label{fig:example}
            \vspace{-0.6cm}
\end{figure}

By leveraging a locally movable permanent magnet, the system achieves precise control over the position of the magnetic micro-robots through a strong, localized magnetic field. Since the DRL model is trained within a grid world generated from vascular model images, it can be effectively translated into real-world scenarios, provided that the micro-robot's position can be accurately mapped from the simulation environment to the physical environment. Fig. \ref{fig:example} compares the performance of the PPO model after its transfer to a real-world setting with its performance in the simulated environment, highlighting the model's robustness and adaptability across both domains.

\subsection{Semi-Autonomous Control}
This paper integrates MR manual control with DRL-based autonomous control to achieve semi-autonomous control of micro-robots, enabling the system to perform tasks such as navigation within a 2D vascular environment.

1) \textbf{MR Manual Control Phase}: In this phase, the operator manually controls the micro-robot by transmitting hand position information via the HoloLens 2 to the micromanipulator. The displacement of the hand is scaled using a factor ($\tau$) \cite{zhang2018self}, and mapped to the micromanipulator's end-effector, as described by the following equation:
\begin{equation}
P_s\left(t\right) = P_s\left(t-1\right) + \tau\left(P_m\left(t\right) - P_m\left(t-1\right)\right)
\label{eq:manual_control}
\end{equation}
where $P_m\left(t\right)$ represents the position of the operator's fingertip at time $t$, and $P_s\left(t\right)$ represents the position of the micromanipulator's end-effector at time $t$.

2) \textbf{DRL Autonomous Control Phase}: During this phase, the navigation policy learned through the PPO algorithm is used to autonomously guide the micro-robot to the target area. The following formula describes how the PPO algorithm samples actions from a set $\{a_1, a_2, \ldots, a_n\}$ based on the probability $\pi_\theta$ of each action, selecting $a_{\mathrm{selected}}$ as the executed action:
\begin{equation}
a_{\mathrm{selected}} \sim \mathrm{Categorical}\left(\pi_\theta\left(a_1\mid s\right),  \ldots, \pi_\theta\left(a_n\mid s\right)\right)
\label{eq:ppo_action}
\end{equation}

By interacting with the real environment, the PPO algorithm generates a motion path for the micro-robot, autonomously navigating it to the critical area.

The system switches between manual and autonomous control modes based on task requirements. For instance, in drug delivery tasks targeting thrombi within narrow blood vessels, the regions containing thrombi are manually marked as critical areas. The DRL policy learned in the simulation environment is then used to quickly and accurately navigate the micro-robot to the critical area. Upon arrival, the control is transferred to human manual operation, allowing the operator to use high-precision manual control for slow, deliberate movements to deliver the drug to the thrombus in the critical area. The switching process is managed by the following formula:
\begin{equation}
\Delta P = |P_s\left(t\right) - P_{\mathrm{critical}}|
\label{eq:switch_control}
\end{equation}
where $\Delta P$ represents the Euclidean distance between the micromanipulator's end-effector position $P_s\left(t\right)$ and the critical position $P_{\mathrm{critical}}$. If $\Delta P \leq \epsilon$, indicating the micro-robot has reached the critical area, the system triggers a mode switch from autonomous control to manual control.

\section{Experiments and Results}

\subsection{Experimental Design}

\subsubsection{Participants and Experimental Setup}

Our user study involved a diverse group of participants that differed in sex, age, familiarity with MR devices, and experience with micromanipulation techniques. Five participants (four males, one female, aged 22 to 27) were involved in this user study. Among them, four had prior experience with MR devices, while only one had priorperience in micromanipulation. To ensure consistency in understanding and execution, verbal instructions were provided to participants who lacked relevant experience before the study began.

\begin{figure}
  \captionsetup{font=footnotesize,labelsep=period}
    \centering
    \includegraphics[width=0.9\linewidth]{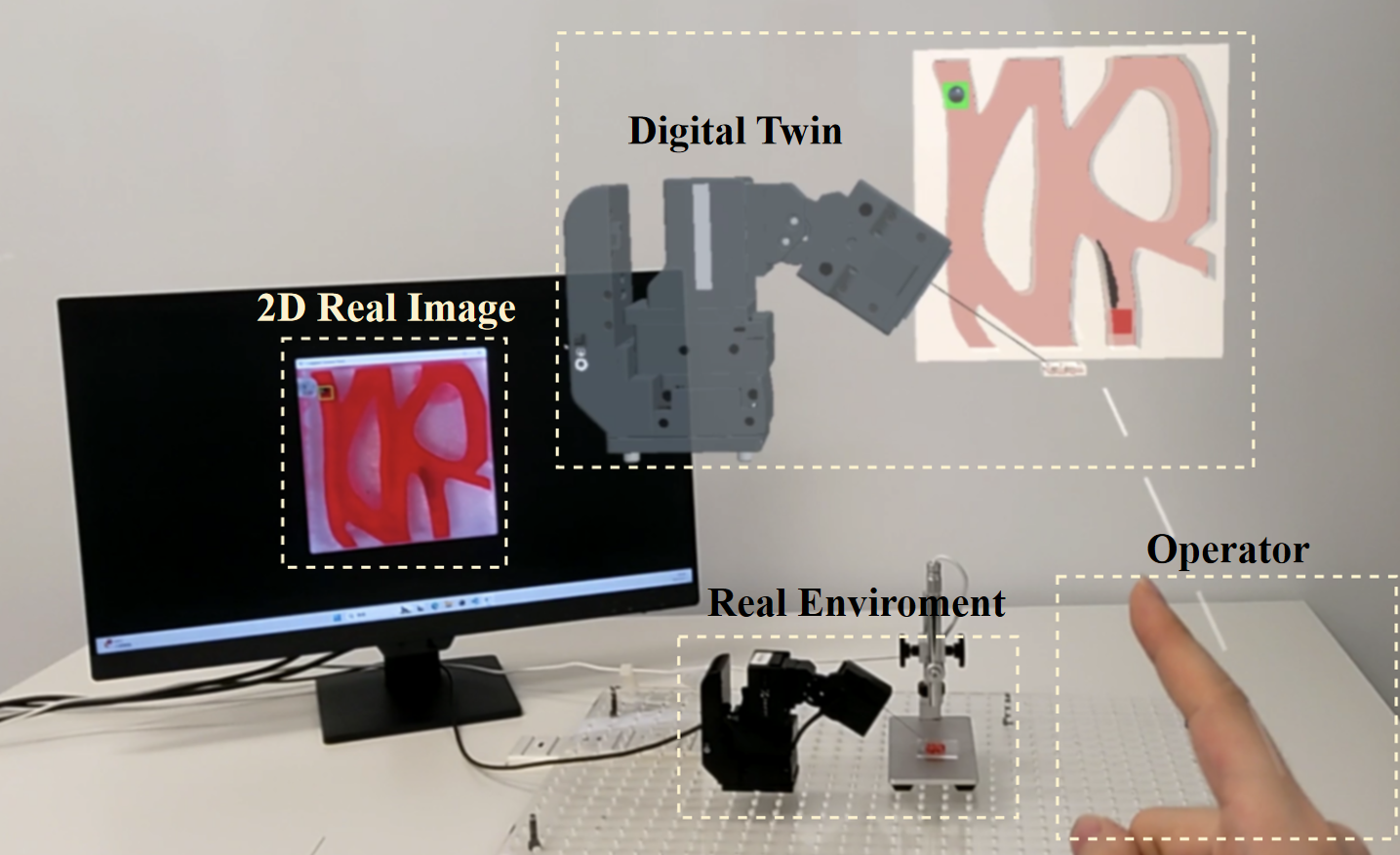} 
    \caption{ The experimental scenario demonstrates teleoperation for navigating a magnetic micro-robot using MR-based manual control. }
    \label{fig:example-setup}
        \vspace{-0.6cm}
\end{figure}

Using a 3D-printed vascular model, we designed a set of controlled experiments to evaluate the DRL-SC framework on magnetic micro-robot navigation control. 
Participants were tasked with using the MR system to control the micromanipulator both with and without DRL-SC. Without DRL-SC, the operator was required to use the MR device to manually control the navigation of the magnetic micro-robot. The experimental setup involved  three distinct 3D-printed microvascular models, each containing simulated thrombus as critical area, as depicted in Fig. \ref{fig:example-setup}.
Before commencing the formal experiments, participants were given equal time in training sessions to familiarize themselves with the experimental protocols. Upon completing the experiments, participants were asked to complete a questionnaire to provide demographic information and to assess their workload using the NASA-TLX (Task Load Index) survey \cite{hart2006nasa}.

\subsubsection{Evaluation Metrics}
\textbf{a) Average Speed (Va)}: This metric is defined as the total path length covered by the magnetic micro-robot from the starting point to the target position divided by the time duration (unit: $\mu$m/s).

\textbf{b) Time of Completion (TOC)}: This metric is defined as the time taken by the magnetic micro-robot to complete the task (unit: seconds).

\textbf{c) Gracefulness (G)}: This metric is used to assess the straightness of the magnetic micro-robot's trajectory, based on the curvature of the path. Gracefulness is defined by the median curvature along the trajectory over a given time period $T_i$, and it is calculated as follows:

\begin{equation}
G = \mathrm{Median}\left(\log_{10}^\kappa{\left(T_i\right)}\right)
\label{eq:gracefulness}
\end{equation}
where $\kappa = \frac{\left|\dot{\lambda}\left(t\right) \times \ddot{\lambda}\left(t\right)\right|}{\left|\dot{\lambda}\left(t\right)\right|^3}$ represents the curvature of the trajectory at any given point, $\lambda(t)$ represents a point in two-dimensional space as a vector, and $\dot{\lambda}\left(t\right)$ and $\ddot{\lambda}\left(t\right)$ denote the instantaneous velocity and acceleration of the magnetic micro-robot, respectively.

\textbf{d) Smoothness (S)}: This metric measures the rate of change of acceleration for the magnetic micro-robot. It is defined as:
\begin{equation}
S = \mathrm{Median}\left(\log_{10}^\phi{\left(T_i\right)}\right)
\label{eq:smoothness}
\end{equation}
where $\phi$ is defined as:
\begin{equation}
\phi = \frac{\sigma^5}{v_p^2} \int_{t-\sigma}^{t} \left|\frac{d^3\lambda\left(t\right)}{dt^3}\right|^2 dt
\label{eq:smoothness_phi}
\end{equation}
where $\sigma$ corresponds to the duration of the time interval, which can be adjusted based on update rates, and $v_p$ denotes the peak speed.

\textbf{e) Collision Count (C)}: This metric counts the number of collisions that occur when the magnetic micro-robot hits the vessel walls.

The NASA-TLX is used to assess the workload of a task by measuring subjective experiences across multiple dimensions. It not only considers the objective difficulty of the task but also takes into account the subjective stress and workload perceived by the individual performing the task. NASA-TLX uses six subscales to measure overall workload: Mental Demand, Physical Demand, Temporal Demand, Effort, Frustration, and Performance. Each dimension is rated on a scale from 0 to 100. In this study, a weighted overall score is used, with weights assigned to each dimension based on pairwise comparisons, and these weights are then applied to the scores of each dimension.

\subsection{Experimental Result}

\begin{figure}
  \captionsetup{font=footnotesize,labelsep=period}
    \centering
    \includegraphics[width=0.8\linewidth]{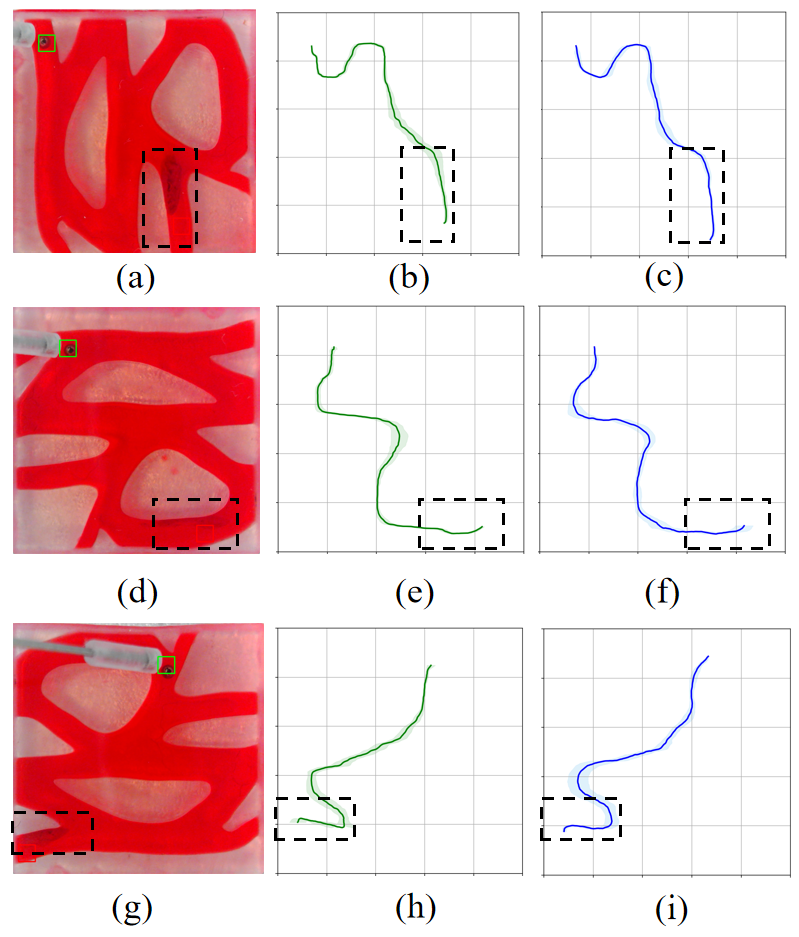} 
    \caption{ (a), (d), (g): Three experimental microvascular models.
(b), (e), (h): Average trajectory using the framework with DRL-SC.
(c), (f), (i): Average trajectory using the framework without DRL-SC.}
    \label{fig:example-real environment}
        \vspace{-0.6cm}
\end{figure}

 Fig. \ref{fig:example-real environment} illustrates the average trajectories of participants using the framework with and without DRL-SC across three 3D-printed vascular models with a simulated thrombus. The trajectories in both modes are very similar, suggesting that participants' navigation closely aligns with the optimal DRL-generated strategies. This implies that the DRL-SC mode aids decision-making and reduces workload. Additionally, the smaller standard deviation of trajectories in the DRL-SC mode indicates more stable navigation strategies and improved efficiency.

\begin{figure}
  \captionsetup{font=footnotesize,labelsep=period}
    \centering
    \includegraphics[width=\linewidth]{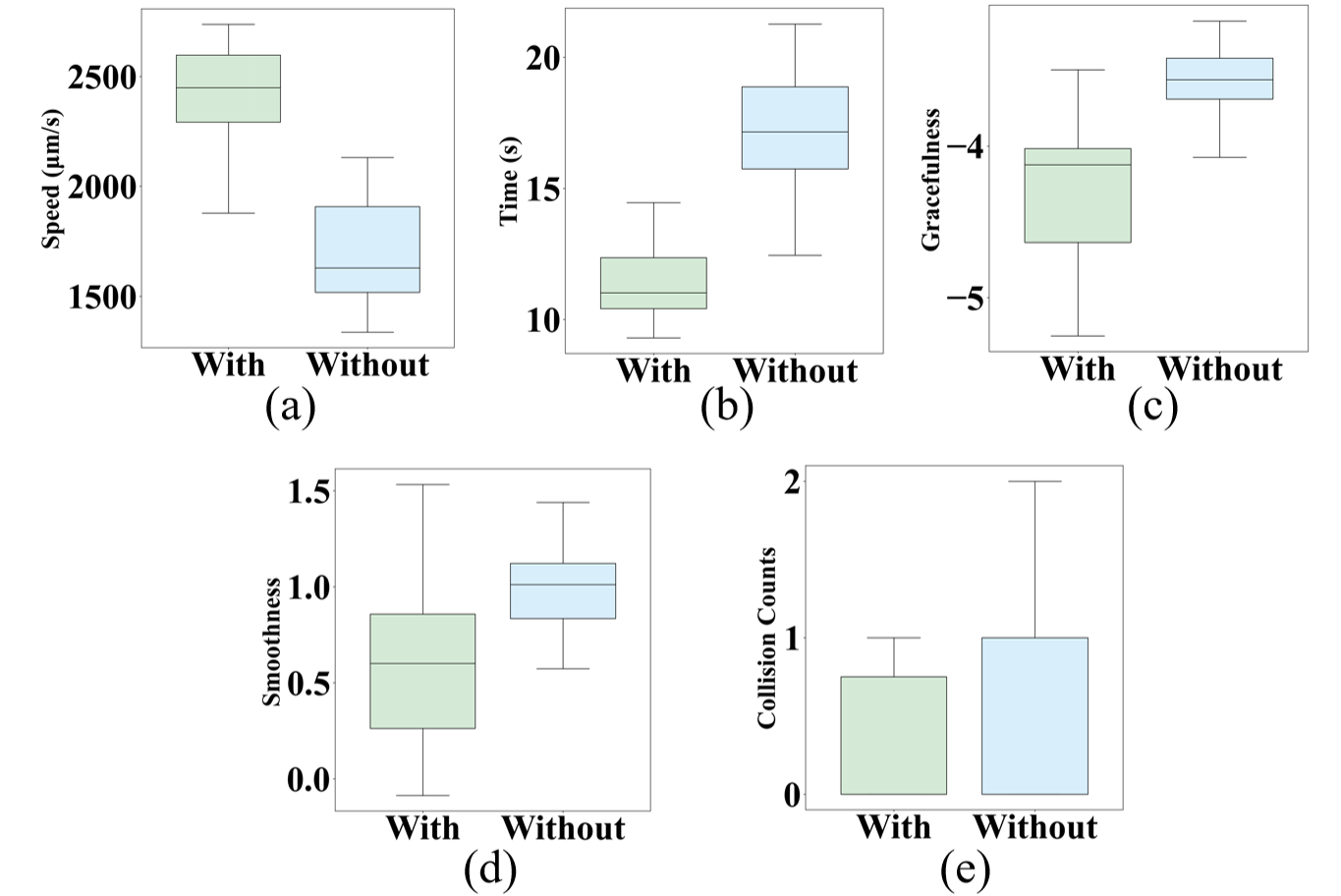} 
    \caption{Box plot comparing the framework with and without DRL-SC in the experiment, using five objective metrics: (a) Average Speed, (b) Completion Time, (c) Gracefulness, (d) Smoothness, and (e) Collision Count.}
    \label{fig:example-Box plot}
\vspace{-0.6cm}
\end{figure}

The results for the five objective metrics are shown in Table \ref{tab:quantitative_results_task2} and Fig. \ref{fig:example-Box plot}. For the manipulation of magnetic micro-robots, a higher average speed and shorter task completion time suggest that participants found the operation easier and had a better understanding of the operational environment, leading to more confidence during operation. 
Lower gracefulness values or higher smoothness values indicate smoother and more stable operation. When using the mode combined with DRL-SC, the average speed increased from 1686.98 $\mu$m/s to 2400.59 $\mu$m/s, and the time required to complete the task was reduced by 33.72\%. This indicates that DRL can help participants make faster decisions to complete the navigation task. The gracefulness metric was -4.45, which is lower than the -3.58 observed in the mode without DRL-SC, and the average collision count per experiment was 0.3, compared to 0.8 in the mode without DRL-SC. These findings suggest that DRL-SC enhances participants' task performance. The only drawback is that DRL-SC did not provide sufficient improvement in the smoothness metric, scoring 0.62, which is lower than the 0.99 observed in the mode without DRL-SC. 


 Further inferential statistical analysis was performed on the five objective metrics. First, the Shapiro-Wilk test was conducted to assess whether the metrics Va, TOC, G, S, and C followed a normal distribution. For the framework without DRL-SC, the p-values for Va, TOC, G, S, and C were 0.061, 0.882, 0.093, 0.490, and 0.0001, respectively. For the framework with DRL-SC, the p-values were 0.026, 0.057, 0.0001, 0.005, and 0.0001, respectively. The Shapiro-Wilk test results indicate that TOC data follows a normal distribution, while the other metrics do not.
 

A t-test was conducted for TOC, yielding a p-value of \(10^{-10}\), which is much less than 0.05, suggesting a statistically significant difference in TOC between conditions with and without DRL-SC. The Mann-Whitney U test was performed for Va, G, S, and C, with p-values of \(3 \times 10^{-10}\), \(6 \times 10^{-9}\), \(1 \times 10^{-10}\), and \(7 \times 10^{-3}\), respectively. These p-values (\(p < 0.05\)) indicate significant statistical differences for these four metrics between conditions with and without DRL-SC.

\begin{table}
    \centering
      \captionsetup{font=footnotesize,labelsep=period}
    \caption{Quantitative results based on the five objective metrics (average values of each experiment).}
    \begin{tabular}{lcc}
        \toprule
        \textbf{Metric} & \textbf{With DRL-SC} & \textbf{Without DRL-SC} \\
        \midrule
        Average Speed (Va) ($\mu$m/s)    & 2400.59 & 1686.98 \\
        Time of Completion (TOC) (s)     & 11.38   & 17.17   \\
        Gracefulness (G)                 & -4.45   & -3.58   \\
        Smoothness (S)                   & 0.62    & 0.99    \\
        Collision Count (C)              & 0.3     & 0.8     \\
        \bottomrule
    \end{tabular}
    \label{tab:quantitative_results_task2}
\end{table}

\begin{figure}
  \captionsetup{font=footnotesize,labelsep=period}
    \centering
    \includegraphics[width=0.9\linewidth]{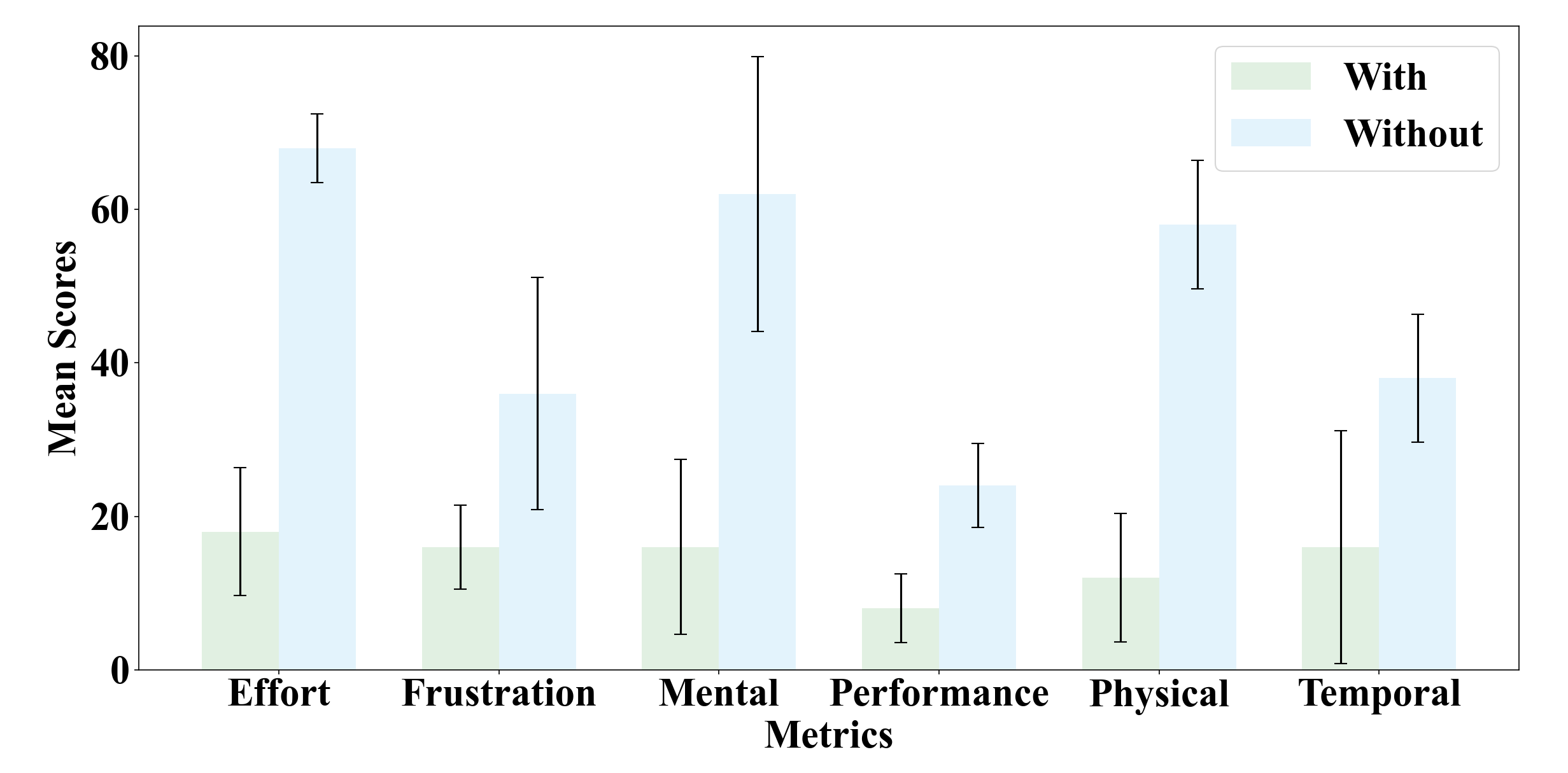} 
    \caption{ Visualization of NASA-TLX results.}
    \label{fig:NASA-TLX results}
       \vspace{-0.6cm}                
\end{figure}

Table \ref{tab:nasa_tlx_task2} and Fig. \ref{fig:NASA-TLX results} show the NASA-TLX scores. For each metric, the scores associated with the framework combined with DRL-SC are lower than those without DRL-SC. This demonstrates that the framework incorporating DRL-SC enhances micromanipulation efficiency and performance, thereby reducing participants' workload.

Experimental results show that the framework enhances user operational efficiency, and accuracy, while reducing workload and burden.
The DRL-SC framework outperformed in navigation performance, with participants achieving higher average speeds, shorter task completion times, and fewer collisions. This suggests that DRL can efficiently identify optimal paths, avoid obstacles, and navigate with precision in narrow vascular models for targeted drug delivery. However, the framework did not significantly improve smoothness, likely due to incomplete transitions between manual and autonomous control, indicating the need for refinement of the semi-autonomous control mechanism.

The DRL-SC framework has considerable limitations. The navigation experiments for magnetic micro-robots were conducted in a 2D environment, and further testing is required to evaluate the framework’s applicability and effectiveness in more complex three-dimensional environments and under real physiological conditions. Additionally, the small number of participants limited the sample size, so expanding the sample size is necessary to validate the generalizability of the current findings.

\begin{table}
  \captionsetup{font=footnotesize,labelsep=period}
    \centering
    \caption{NASA-TLX scores.}
    \begin{tabular}{lcc}
        \toprule
        \textbf{Metric} & \textbf{With DRL-SC} & \textbf{Without DRL-SC} \\
        \midrule
        Weighted Score      & 14.0  & 48.4 \\
        Effort              & 18    & 68   \\
        Frustration Level   & 16    & 36   \\
        Mental Demand       & 16    & 62   \\
        Performance         & 8     & 24   \\
        Physical Demand     & 12    & 58   \\
        Temporal Demand     & 16    & 38   \\
        \bottomrule
    \end{tabular}
    \label{tab:nasa_tlx_task2}
    \vspace{-0.2cm}
\end{table}

\section{Conclusions}

This study introduces a novel framework that integrates DRL-SC and MR technology for navigating magnetic micro-robots in simulated microvascular systems. Experimental results demonstrate that this framework improves intuitive user control, operational efficiency, and accuracy while reducing workload and cognitive burden during micro-robot control. The semi-autonomous control approach has the potential to expand the application of magnetic micro-robots in biomedical fields, particularly for tasks requiring precise, adaptive, safe, and intelligent non-invasive interventions.
Future work will focus on extending the DRL-SC framework to 3D microfluidic systems with more complex microvascular structures. Additionally, its performance will be evaluated in biofluidic environments to further validate its effectiveness and adaptability.


\section*{Acknowledge}

Yudong Mao would like to acknowledge Q. Cong and J. Lin's support and advice during his MSc study.

This research was conducted in accordance with ethical guidelines and was approved by the Imperial College London Research Ethics Committee (Approval ID: 7313273).

\bibliographystyle{IEEEtran}
\bibliography{references}

\begin{thebibliography}{10}
\providecommand{\url}[1]{#1}
\csname url@samestyle\endcsname
\providecommand{\newblock}{\relax}
\providecommand{\bibinfo}[2]{#2}
\providecommand{\BIBentrySTDinterwordspacing}{\spaceskip=0pt\relax}
\providecommand{\BIBentryALTinterwordstretchfactor}{4}
\providecommand{\BIBentryALTinterwordspacing}{\spaceskip=\fontdimen2\font plus
\BIBentryALTinterwordstretchfactor\fontdimen3\font minus
  \fontdimen4\font\relax}
\providecommand{\BIBforeignlanguage}[2]{{%
\expandafter\ifx\csname l@#1\endcsname\relax
\typeout{** WARNING: IEEEtran.bst: No hyphenation pattern has been}%
\typeout{** loaded for the language `#1'. Using the pattern for}%
\typeout{** the default language instead.}%
\else
\language=\csname l@#1\endcsname
\fi
#2}}
\providecommand{\BIBdecl}{\relax}
\BIBdecl

\bibitem{luo2018micro}
M.~Luo, Y.~Feng, T.~Wang, and J.~Guan, ``Micro-/nanorobots at work in active
  drug delivery,'' \emph{Advanced Functional Materials}, vol.~28, no.~25, p.
  1706100, 2018.

\bibitem{lin2024magnetic}
J.~Lin, Q.~Cong, and D.~Zhang, ``Magnetic microrobots for in vivo cargo
  delivery: a review,'' \emph{Micromachines}, vol.~15, no.~5, p. 664, 2024.

\bibitem{vikram2016targeted}
A.~Vikram~Singh and M.~Sitti, ``Targeted drug delivery and imaging using mobile
  milli/microrobots: A promising future towards theranostic pharmaceutical
  design,'' \emph{Current pharmaceutical design}, vol.~22, no.~11, pp.
  1418--1428, 2016.

\bibitem{zhang2023advanced}
D.~Zhang, T.~E. Gorochowski, L.~Marucci, H.-T. Lee, B.~Gil, B.~Li, S.~Hauert,
  and E.~Yeatman, ``Advanced medical micro-robotics for early diagnosis and
  therapeutic interventions,'' \emph{Frontiers in Robotics and AI}, vol.~9, p.
  1086043, 2023.

\bibitem{wang2021trends}
B.~Wang, K.~Kostarelos, B.~J. Nelson, and L.~Zhang, ``Trends in
  micro-/nanorobotics: materials development, actuation, localization, and
  system integration for biomedical applications,'' \emph{Advanced Materials},
  vol.~33, no.~1, p. 2002047, 2021.

\bibitem{zhang2022fabrication}
D.~Zhang, Y.~Ren, A.~Barbot, F.~Seichepine, B.~Lo, Z.-C. Ma, and G.-Z. Yang,
  ``Fabrication and optical manipulation of micro-robots for biomedical
  applications,'' \emph{Matter}, vol.~5, no.~10, pp. 3135--3160, 2022.

\bibitem{wang2023microrobots}
Y.~Wang \emph{et~al.}, ``Microrobots for targeted delivery and therapy in
  digestive system,'' \emph{ACS Nano}, vol.~17, no.~1, pp. 27--50, 2023.

\bibitem{wang2021endoscopy}
B.~Wang \emph{et~al.}, ``Endoscopy-assisted magnetic navigation of biohybrid
  soft microrobots with rapid endoluminal delivery and imaging,'' \emph{Science
  Robotics}, vol.~6, no.~56, p. eabd2813, 2021.

\bibitem{abbes2019permanent}
M.~Abbes, K.~Belharet, H.~Mekki, and G.~Poisson, ``Permanent magnets based
  actuator for microrobots navigation,'' in \emph{2019 IEEE/RSJ International
  Conference on Intelligent Robots and Systems (IROS)}, 2019, pp. 7062--7067.

\bibitem{nadour2023cochlerob}
H.~Nadour, A.~B. Grayeli, G.~Poisson, and K.~Belharet, ``Cochlerob:
  Parallel-serial robot to position a magnetic actuator around a patient’s
  head for intracochlear microrobot navigation,'' \emph{Sensors}, vol.~23,
  no.~6, p. 2973, 2023.

\bibitem{lee2018capsule}
S.~Lee, S.~Kim, S.~Kim, J.-Y. Kim, C.~Moon, B.~J. Nelson, and H.~Choi, ``A
  capsule-type microrobot with pick-and-drop motion for targeted drug and cell
  delivery,'' \emph{Advanced Healthcare Materials}, vol.~7, no.~9, p. 1700985,
  2018.

\bibitem{lee2020needle}
S.~Lee \emph{et~al.}, ``A needle-type microrobot for targeted drug delivery by
  affixing to a microtissue,'' \emph{Advanced Healthcare Materials}, vol.~9,
  no.~7, p. 1901697, 2020.

\bibitem{jeong2020acoustic}
J.~Jeong, D.~Jang, D.~Kim, D.~Lee, and S.~K. Chung, ``Acoustic bubble-based
  drug manipulation: Carrying, releasing and penetrating for targeted drug
  delivery using an electromagnetically actuated microrobot,'' \emph{Sensors
  and Actuators A: Physical}, vol. 306, p. 111973, 2020.

\bibitem{gong2024magnetic}
D.~Gong, N.~Celi, D.~Zhang, and J.~Cai, ``Magnetic biohybrid microrobot
  multimers based on chlorella cells for enhanced target,'' 2024.

\bibitem{abdelaziz2022electromagnetic}
M.~Abdelaziz and M.~Habib, ``Electromagnetic actuation for a micro/nano robot
  in a three-dimensional environment,'' \emph{Micromachines}, vol.~13, no.~11,
  p. 2028, 2022.

\bibitem{erni2013comparison}
S.~Erni, S.~Schurle, A.~Fakhraee, B.~E. Kratochvil, and B.~J. Nelson,
  ``Comparison, optimization, and limitations of magnetic manipulation
  systems,'' \emph{Journal of Micro-Bio Robotics}, vol.~8, pp. 107--120, 2013.

\bibitem{lee2021real}
J.~Lee, X.~Zhang, C.~H. Park, and M.~J. Kim, ``Real-time teleoperation of
  magnetic force-driven microrobots with 3d haptic force feedback for
  micro-navigation and micro-transportation,'' \emph{IEEE Robotics and
  Automation Letters}, vol.~6, no.~2, pp. 1769--1776, Apr. 2021.

\bibitem{Jang_2019}
D.~Jang, J.~Jeong, H.~Song, and S.~K. Chung, ``Targeted drug delivery
  technology using untethered microrobots: a review,'' \emph{Journal of
  Micromechanics and Microengineering}, vol.~29, no.~5, p. 053002, mar 2019.

\bibitem{mehrtash2011bilateral}
M.~Mehrtash, N.~Tsuda, and M.~B. Khamesee, ``Bilateral macro--micro
  teleoperation using magnetic levitation,'' \emph{IEEE/ASME Transactions on
  Mechatronics}, vol.~16, no.~3, pp. 459--469, 2011.

\bibitem{khatib2020teleoperation}
E.~Al~Khatib, X.~Zhang, M.~J. Kim \emph{et~al.}, ``Teleoperation control scheme
  for magnetically actuated microrobots with haptic guidance,'' \emph{Journal
  of Micro-Bio Robotics}, vol.~16, no. 3-4, pp. 161--171, 2020.

\bibitem{jiang2024digital}
P.~Jiang and D.~Zhang, ``A digital twin-driven immersive teleoperation
  framework for robot-assisted microsurgery,'' in \emph{2024 IEEE/RSJ
  International Conference on Intelligent Robots and Systems (IROS)}.\hskip 1em
  plus 0.5em minus 0.4em\relax IEEE, 2024, pp. 13\,495--13\,501.

\bibitem{hoenig2015mixed}
W.~Hoenig, C.~Milanes, L.~Scaria, T.~Phan, M.~Bolas, and N.~Ayanian, ``Mixed
  reality for robotics,'' in \emph{2015 IEEE/RSJ International Conference on
  Intelligent Robots and Systems (IROS)}.\hskip 1em plus 0.5em minus
  0.4em\relax IEEE, 2015, pp. 5382--5387.

\bibitem{fan2023digital}
W.~Fan, X.~Guo, E.~Feng, J.~Lin, Y.~Wang, J.~Liang, M.~Garrad, J.~Rossiter,
  Z.~Zhang, N.~Lepora \emph{et~al.}, ``Digital twin-driven mixed reality
  framework for immersive teleoperation with haptic rendering,'' \emph{IEEE
  Robotics and Automation Letters}, vol.~8, no.~12, pp. 8494--8501, 2023.

\bibitem{zhang2024hubotverse}
D.~Zhang, Z.~Wu, J.~Zheng, Y.~Li, Z.~Dong, and J.~Lin, ``Hubotverse: Toward
  internet of human and intelligent robotic things with a digital twin-based
  mixed reality framework,'' \emph{IEEE Robotics \& Automation Magazine}, 2024.

\bibitem{chowdhury2024virtual}
A.~M. M.~B. Chowdhury, S.~A. Abbasi, N.~L. Gharamaleki, J.~Kim, and H.~Choi,
  ``Virtual reality-enabled intuitive magnetic manipulation of microrobots,''
  \emph{Advanced Intelligent Systems}, vol.~6, no.~7, p. 2300793, Mar. 2024.

\bibitem{salehi2024intelligent}
A.~Salehi, S.~Hosseinpour, N.~Tabatabaei, M.~S. Firouz, and T.~Yu,
  ``Intelligent navigation of a magnetic microrobot with model-free deep
  reinforcement learning in a real-world environment,'' \emph{Micromachines},
  vol.~15, no.~1, p. 112, Jan. 2024.

\bibitem{li2024automated}
Y.~Li, Y.~Huo, X.~Chu, and L.~Yang, ``Automated magnetic microrobot control:
  From mathematical modeling to machine learning,'' \emph{Mathematics},
  vol.~12, p. 2180, 2024, [Online]. Available:
  https://doi.org/10.3390/math12142180.

\bibitem{das2024multifaceted}
T.~Das and S.~Sultana, ``Multifaceted applications of micro/nanorobots in
  pharmaceutical drug delivery systems: a comprehensive review,'' \emph{Future
  Journal of Pharmaceutical Sciences}, vol.~10, p.~2, 2024, [Online].
  Available: https://doi.org/10.1186/s43094-023-00577-y.

\bibitem{wang2024deep}
H.~Wang, Y.~Qiu, Y.~Hou, Q.~Shi, H.-W. Huang, Q.~Huang, and T.~Fukuda, ``Deep
  reinforcement learning-based collision-free navigation for magnetic helical
  microrobots in dynamic environments,'' \emph{IEEE Transactions on Automation
  Science and Engineering}, 2024.

\bibitem{abbasi2024autonomous}
S.~A. Abbasi \emph{et~al.}, ``Autonomous 3d positional control of a magnetic
  microrobot using reinforcement learning,'' \emph{Nature Machine
  Intelligence}, vol.~6, pp. 92--105, Jan. 2024.

\bibitem{behrens2022smart}
M.~R. Behrens and W.~C. Ruder, ``Smart magnetic microrobots learn to swim with
  deep reinforcement learning,'' \emph{Advanced Intelligent Systems}, vol.~4,
  no.~10, p. 2200023, 2022.

\bibitem{zhang2022human}
D.~Zhang, Z.~Wu, J.~Chen, R.~Zhu, A.~Munawar, B.~Xiao, Y.~Guan, H.~Su, W.~Hong,
  Y.~Guo \emph{et~al.}, ``Human-robot shared control for surgical robot based
  on context-aware sim-to-real adaptation,'' in \emph{2022 International
  conference on robotics and automation (ICRA)}.\hskip 1em plus 0.5em minus
  0.4em\relax IEEE, 2022, pp. 7694--7700.

\bibitem{9341383}
J.~Chen, D.~Zhang, A.~Munawar, R.~Zhu, B.~Lo, G.~S. Fischer, and G.-Z. Yang,
  ``Supervised semi-autonomous control for surgical robot based on bayesian
  optimization,'' in \emph{2020 IEEE/RSJ International Conference on
  Intelligent Robots and Systems (IROS)}, 2020, pp. 2943--2949.

\bibitem{zhang2020microsurgical}
D.~Zhang, J.~Chen, W.~Li, D.~Bautista~Salinas, and G.-Z. Yang, ``A
  microsurgical robot research platform for robot-assisted microsurgery
  research and training,'' \emph{International journal of computer assisted
  radiology and surgery}, vol.~15, pp. 15--25, 2020.

\bibitem{jiang2024adaptive}
P.~Jiang, W.~Li, Y.~Li, and D.~Zhang, ``Adaptive motion scaling for
  robot-assisted microsurgery based on hybrid offline reinforcement learning
  and damping control,'' in \emph{2024 IEEE International Conference on
  Robotics and Automation (ICRA)}.\hskip 1em plus 0.5em minus 0.4em\relax IEEE,
  2024, pp. 8216--8222.

\bibitem{mnih2016asynchronous}
V.~Mnih, A.~P. Badia, M.~Mirza, A.~Graves, T.~Lillicrap, T.~Harley, D.~Silver,
  and K.~Kavukcuoglu, ``Asynchronous methods for deep reinforcement learning,''
  in \emph{International conference on machine learning}.\hskip 1em plus 0.5em
  minus 0.4em\relax PmLR, 2016, pp. 1928--1937.

\bibitem{schulman2017proximalpolicyoptimizationalgorithms}
J.~Schulman, F.~Wolski, P.~Dhariwal, A.~Radford, and O.~Klimov, ``Proximal
  policy optimization algorithms,'' 2017.

\bibitem{zhang2018self}
D.~Zhang, B.~Xiao, B.~Huang, L.~Zhang, J.~Liu, and G.-Z. Yang, ``A
  self-adaptive motion scaling framework for surgical robot remote control,''
  \emph{IEEE Robotics and Automation Letters}, vol.~4, no.~2, pp. 359--366,
  2018.

\bibitem{hart2006nasa}
S.~G. Hart, ``Nasa-task load index (nasa-tlx); 20 years later,'' in
  \emph{Proceedings of the human factors and ergonomics society annual
  meeting}, vol.~50, no.~9.\hskip 1em plus 0.5em minus 0.4em\relax Sage
  publications Sage CA: Los Angeles, CA, 2006, pp. 904--908.

\end{thebibliography}

\end{document}